\documentclass[10pt,twocolumn,letterpaper]{article}

\usepackage{cvpr}
\usepackage{times}
\usepackage{epsfig}
\usepackage{graphicx}
\usepackage{amsmath}
\usepackage{amssymb}
\usepackage{booktabs}
\usepackage{adjustbox}
\usepackage{multirow}
\usepackage{float}
\usepackage{subcaption}
\usepackage{xcolor}
\graphicspath{{figures/}}

\usepackage[pagebackref=true,breaklinks=true,letterpaper=true,colorlinks,bookmarks=false]{hyperref}

\cvprfinalcopy 

\def\cvprPaperID{18} 

\ifcvprfinal\fi
\begin{document}

\title{Offline Signature Verification on Real-World Documents}

\author{Deniz Engin$^{1}$\thanks{indicates equal contribution} \qquad
Alperen Kantarc{\i}$^{2}$\footnotemark[1] \qquad
Se{\c{c}}il Arslan$^{1}$ \qquad
Haz{\i}m Kemal Ekenel$^{2}$\\ \\
$^{1}$Yap{\i} Kredi Technology\\
$^{2}$Istanbul Technical University\\
{\tt\small \{deniz.engin,secil.arslan\}@ykteknoloji.com.tr \qquad \{kantarcia,ekenel\}@itu.edu.tr}
}

\maketitle
\begin{abstract}
Research on offline signature verification has explored a large variety of methods on multiple signature datasets, which are collected under controlled conditions. However, these datasets may not fully reflect the characteristics of the signatures in some practical use cases. Real-world signatures extracted from the formal documents may contain different types of occlusions, for example, stamps, company seals, ruling lines, and signature boxes. Moreover, they may have very high intra-class variations, where even genuine signatures resemble forgeries. In this paper, we address a real-world writer independent offline signature verification problem, in which, a bank's customers' transaction request documents that contain their occluded signatures are compared with their clean reference signatures. Our proposed method consists of two main components, a stamp cleaning method based on CycleGAN and signature representation based on CNNs. We extensively evaluate different verification setups, fine-tuning strategies, and signature representation approaches to have a thorough analysis of the problem. Moreover, we conduct a human evaluation to show the challenging nature of the problem. We run experiments both on our custom dataset, as well as on the publicly available Tobacco-800 dataset. The experimental results validate the difficulty of offline signature verification on real-world documents. However, by employing the stamp cleaning process, we improve the signature verification performance significantly.


\end{abstract}

\section{Introduction}

Handwritten signatures are one of the oldest and most widely used biometric authentication techniques in administrative and financial institutions due to its simplicity and uniqueness~\cite{onlineofflinesurvey}. As technology progresses, authentication methods have also evolved. Handwritten signatures are now categorized as online signatures and offline signatures. Online signatures have much more distinct features than offline signatures; therefore, they are easier to verify~\cite{analyzedeepcnn}. However, capturing online signatures is expensive, and digital systems prefer different authentication methods, such as passwords or personal authentication questions. On the other hand, offline signatures are easy to capture but hard to verify due to the limited amount of features they contain and uncontrolled environmental acquisition conditions.

\begin{figure}[t!]
  \centering
    \includegraphics[width=1\linewidth]{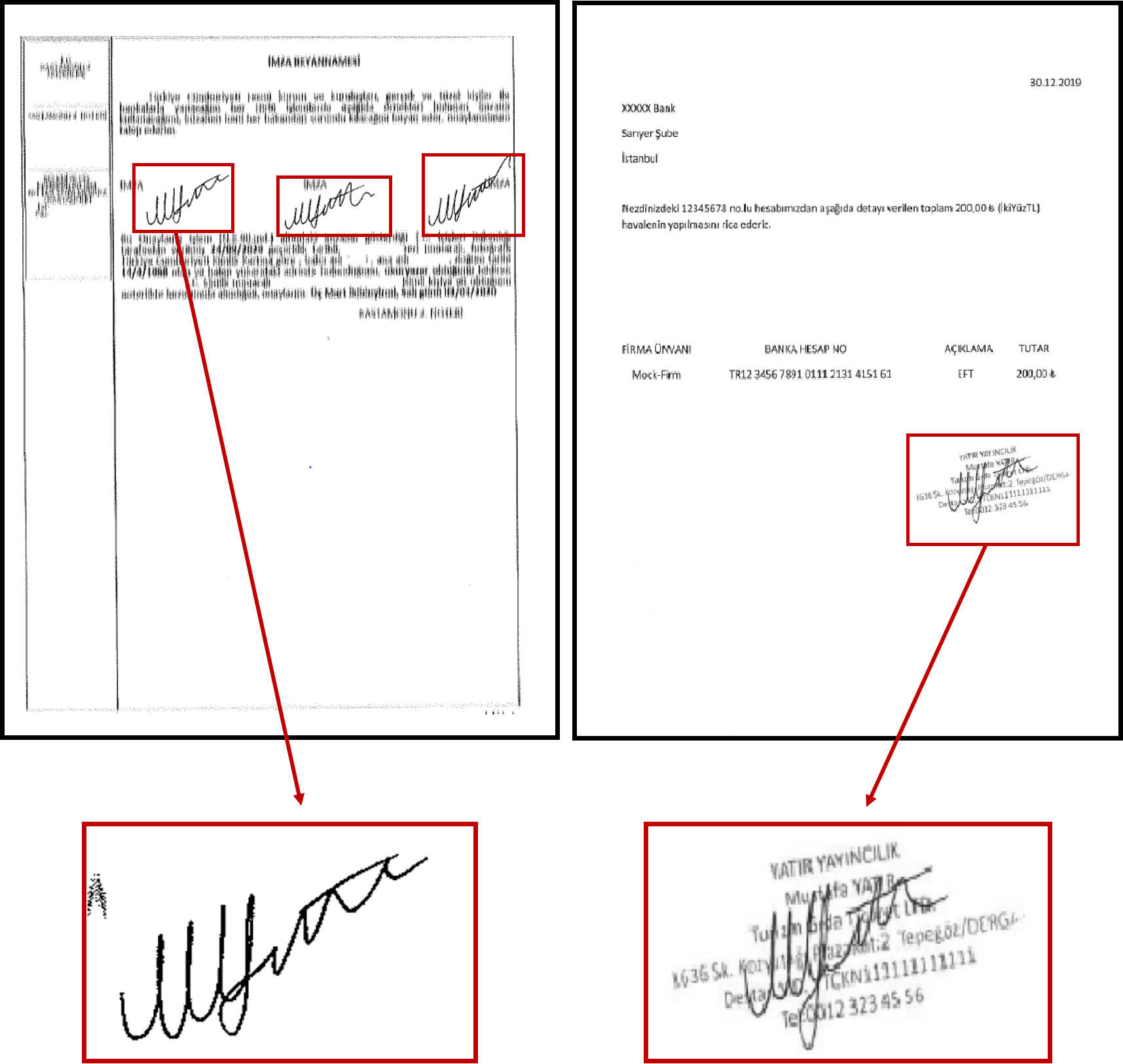} 
  \caption{Example of verifying a signature extracted from a signature declaration document (left) and the stamped signature extracted from an order document  (right).}
  \label{fig:example_sirku_talimat}
\end{figure}

Offline signature verification task has been a challenge for computer vision research and many different approaches have been proposed to perform the task more accurately~\cite{onlineofflinesurvey}. Evaluations of these approaches have been conducted on publicly available datasets such as GPDS-960~\cite{gpds960}, GPDS-4000~\cite{gpds4000synthetic}, MCYT~\cite{mcyt75}, and CEDAR~\cite{cedar}. All of these datasets contain genuine signatures of the users with random and skilled forgeries that try to imitate the genuine signature. The collection of the signatures is completed in either single or multiple sessions. People tend to sign very similar signatures when they sign one after another, similar to the one session acquisition, however, signatures differ very much when signatures are collected over time. In real-world applications, signatures of a person can be varied considerably, because people sign a lot of documents in their daily lives, and it is unlikely to sign exactly the same every time. Therefore, datasets, that acquire signatures in a short period, do not capture the high intra-class variety of a person's signature.

In the literature, both writer dependent and writer independent signature verification methods have been proposed. However, in a real-world signature verification setting user enrollment is very frequent. On account of this reason, writer dependent methods are not feasible to apply. 
In the writer independent methods, the subjects used for training and testing are different, so no person specific features can be utilized. Writer independent methods try to learn efficient representations of the signatures to distinguish each person, but creating a universal discriminative representation of a signature is challenging and no particular feature extraction method has been found to solve this problem \cite{GPDSDeepCnn}. 

In this paper, we focus on offline signature verification in the banking process as one of the real-world application scenarios. In the banks, the customers from enterprise and commercial segments send their banking transaction orders with mainly petition-based documents. These documents are received by the central operation unit of the banks from their fax, scanner, and e-mail channels. The operators are responsible for checking the signature, whether it is the same with the one, which is seen on the signature declaration document of the same customer. This task is illustrated in Figure~\ref{fig:example_sirku_talimat}. Due to the requirement of a significant manual workforce, we aim to automatize this process for the documents of the companies that have exactly one authorized employee to sign the documents. It is measured that such types of customers send around 90,000 pages of banking order documents per month in the medium-size banks. The signature verification task with a manual workforce requires approximately 233 person-hours to process these documents. Hence, employing an automatic offline signature verification system provides significant resource efficiency to the central operation unit of banks.

In this work, we collect bank order and signature declaration documents of the customer's\footnote{Please note that due to data confidentiality, we cannot publish samples from our real-world dataset. Therefore, to visualize our real-world signature verification problem, imitations of signatures, rubber-stamps, and document images are provided from our dataset. 
}. The location of signatures on these documents are annotated manually. This way, we create a real-world signature dataset. Signatures on order documents can be rubber-stamped or unstamped. Therefore, we need a stamp cleaning method to obtain more clear signatures before the verification process. Inspired from image-to-image translation works in the literature, we utilize the CycleGAN~\cite{cyclegan} for stamp cleaning.  We generate two datasets from the created signature dataset, one for representation learning and the other to run verification tests. These two subsets contain signatures from different individuals. Thus, we train a deep feature extraction network on a completely different set of users than the ones in the test set to have a writer independent feature extractor. 

Please note that we cannot make our confidential customer signature dataset of the bank publicly available due to the General Data Protection Regulation (GDPR). Furthermore, we cannot use publicly available signature verification datasets, such as GPDS-960~\cite{gpds960}, GPDS-4000~\cite{gpds4000synthetic}, MCYT~\cite{mcyt75}, and CEDAR~\cite{cedar}, because our problem differs from the one presented by them regarding data collection and application purpose. Therefore, we also prepare another real-world signature verification setup using the publicly available Tobacco-800 dataset~\cite{tobacco800,tobacco_website} and conduct experiments on it. In order to promote signature verification research on real-world documents, we publish the generated training, validation, and verification splits that we use in this benchmark\footnote{https://github.com/Alpkant/Offline-Signature-Verification-on-Real-World-Documents}. \\

Our main contributions can be summarised as follows:

\begin{itemize}
    \item We present a comprehensive study on offline signature verification on real-world documents. For this purpose, we both create a custom offline signature verification dataset and a real-world signature verification setup using the publicly available Tobacco-800 dataset.   
    
    \item We extensively analyze different verification setups, fine-tuning strategies, and signature representation approaches. Moreover, we conduct a human evaluation to show the challenging nature of the problem.
 
    \item We formulate the stamp removal task as an unpaired image-to-image translation problem and propose a CycleGAN-based stamp removal method. With the proposed framework, we achieve a significant reduction in the equal error rate.  
\end{itemize}

The remainder of the paper is organized as follows. In Section 2, we review the related work. The proposed method is explained in Section 3. Experimental setups and the corresponding results are presented and discussed in Section 4. Finally, Section 5 concludes the paper.


\section{Related Work}


\textbf{Noise Cleaning.} Signatures on the complex documents often overlap with different parts of the documents, such as stamps, ruling lines, printed and handwritten texts, which are called noise in general. Removal of these parts can be seen as a segmentation problem since segmented parts can be removed to extract a clean signature. \cite{dstarstampsegmentation} proposed a fully convolutional stamp segmentation network to detect different kinds of stamps in the documents. Stamps change a lot between companies and countries; therefore, network training for the specific dataset is essential. Their proposed network has been trained with pixel-level stamp annotations; however, creating a pixel level stamp annotation for real-world documents is not feasible. On account of this, we utilize a noise cleaning method, which does not require pixel-level annotations, and it is trained in an unpaired manner. 

In \cite{sharma2018learning}, a CycleGAN \cite{cyclegan} based scanning artifact removal deep network is proposed to clean documents from a variety of noises, \eg, watermark, background noise, and blur. They train their network on four different datasets for four different noise types, however, these datasets are synthetically created. Our proposed noise-cleaning network has been trained on real-world documents. Moreover, we do not constrain our network to a limited number of noise types or degradations. 
For example, printed and handwritten texts or stamps are also seen as noise along with the other noise types for our network. \\ 

\textbf{Signature Verification.} Like all other computer vision problems, handcrafted features have been widely used in the signature verification. \cite{yilmaz2011handcrafted} built a support vector machine (SVM) classifier on top of combined local binary patterns (LBP) and histogram of oriented gradients (HOG) features. This approach achieved the highest score in ICDAR SigWiComp challenge both in 2013 \cite{icdar2013} and 2015 \cite{icdar2015}. Instead of searching good handcrafted features, deep convolutional neural networks have been utilized  to learn feature representations from raw data \cite{signet, twochannel, deephsv}. In \cite{analyzedeepcnn}, the authors investigated the feature representations of the deep learning models specifically for the signatures. Analysis of the features showed that deep learning models could successfully create good representations of the signatures and able to discriminate the genuine signatures. Also, \cite{GPDSDeepCnn} created a writer independent deep neural convolutional network to prove that learned feature space not only generalizes to unseen users in a dataset but also to the users from other datasets. This is also a good indicator of the applicability of the deep convolutional neural networks to the real-world signature verification task. \cite{inverseverification} proposed a multiple stream verification network, which uses original and inverse signatures. They claim that their network focuses more on the signature strokes when original and inverse signatures are used together with their inverse streams and multi-path attention modules.

\section{Proposed Method}


Our proposed system includes two main steps as stamp cleaning and representation learning. In the system, after stamp cleaning, signature representations are extracted. Then similarity between two signature representations is measured and compared to a general threshold to determine whether the signatures belong to the same person or not. In the following subsections, we explain these processes.


\subsection{Stamp Cleaning}

Signatures on the real-world documents might be stamped, which degrades the verification process. In our dataset, the target signatures generally include a stamp. Thus, a conversion between stamped and unstamped signatures is a critical process for signature verification. For this reason, a stamp cleaning method is necessary. The requirement of an unsupervised method is the primary constraint for the stamp cleaning method due to the difficulty of collecting a large number of stamped and unstamped pairs of signatures from the same users in real-world documents. This limitation motivates us to utilize CycleGAN~\cite{cyclegan} to perform unpaired image-to-image translation.

\begin{figure}[t!]
  \centering
  \includegraphics[width=0.95\linewidth]{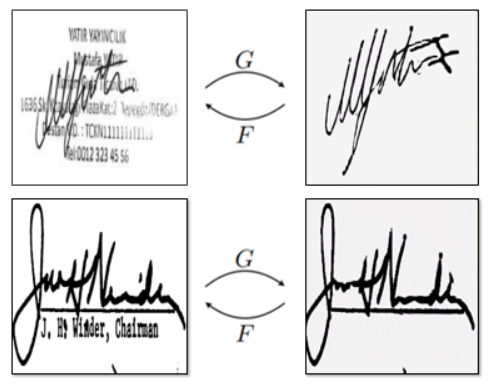}
  \caption{Examples of stamped and cleaned signatures from our dataset (above) and Tobacco-800 (below).}
  \label{fig:noise_clean}
\end{figure}

We collect a dataset by using the extracted signatures from the documents. There are 1287 signatures extracted from signature declaration documents which are clean, whereas 3607 signatures extracted from the order documents contain stamps. Our aim is to learn the conversion between stamped signatures, X, and unstamped ones, Y. For this purpose, two mapping functions $G: X \xrightarrow[]{} Y$ and $F: Y \xrightarrow[]{} X$ are defined.

The adversarial loss for mapping function $G: X \xrightarrow[]{} Y$ is given in Equation~\ref{eq:adversarial_loss}. Adversarial loss for mapping function $F: Y \xrightarrow[]{} X$ is also similar to this adversarial loss. 
\begin{equation}\begin{aligned}
\mathcal{L}_{\mathrm{GAN}}\left(G, D_{Y}, X, Y\right) &=\mathbb{E}_{y \sim p_{\text {data }}(y)}\left[\log D_{Y}(y)\right] \\
&+\mathbb{E}_{x \sim p_{\text {eun }}(x)}\left[\log \left(1-D_{Y}(G(x))\right]\right.
\end{aligned}
\label{eq:adversarial_loss}
\end{equation}
 As an improvement to adversarial loss, cycle consistency loss has been proposed in CycleGAN to compare generated images with input images using the cyclic process. In cycle consistency loss described in Equation~\ref{eq:cycle_loss}, the L1 norm is employed to calculate the loss between generated inputs and original inputs.
\begin{equation}\begin{aligned}
\mathcal{L}_{\text {cyc }}(G, F) &=\mathbb{E}_{x \sim p_{\text {dea }}(x)}\left[\|F(G(x))-x\|_{1}\right] \\
&+\mathbb{E}_{y \sim p_{\text {dea }}(y)}\left[\|G(F(y))-y\|_{1}\right]
\end{aligned}\label{eq:cycle_loss}
\end{equation}
The full objective of CycleGAN, which consists of adversarial losses in two ways and cycle consistency loss, is given in Equation~\ref{eq:full_objective}.
\begin{equation}\begin{aligned}
\mathcal{L}\left(G, F, D_{X}, D_{Y}\right) &=\mathcal{L}_{\text {GAN }}\left(G, D_{Y}, X, Y\right) \\
&+\mathcal{L}_{\text {GNN }}\left(F, D_{X}, Y, X\right) \\
&+\lambda \mathcal{L}_{\text {cyc }}(G, F)
\end{aligned}\label{eq:full_objective}
\end{equation}

The sample inputs and outputs of our cleaning process can be seen in Figure~\ref{fig:noise_clean}. Our trained model is able to remove texts successfully on images in both datasets.

\subsection{Representation Learning}

Writer dependent signature verification models are not feasible for real-world signature verification scenarios where user enrollment is very frequent. Therefore, we should learn writer independent signature representations to verify signatures. For this purpose, we benefit from well-known, successful architectures, namely, VGG-16 \cite{vgg16} and ResNet-50~\cite{resnet50}, and their pre-trained models on ImageNet~\cite{ImageNet}. 
For each dataset, we fine-tune these networks' models with signatures of the users in the training set. In the verification test set, we have signatures of the users that our networks have never seen before. 
For each network architecture, we fine-tune three models with different settings: raw signature images, cleaned signature images, and inverse signature images. By changing the input image type, we explore the effect of the cleaned and inverse signature images.

\subsection{Verification} 

\begin{figure}[t!]
  \centering
  \includegraphics[width=1\linewidth]{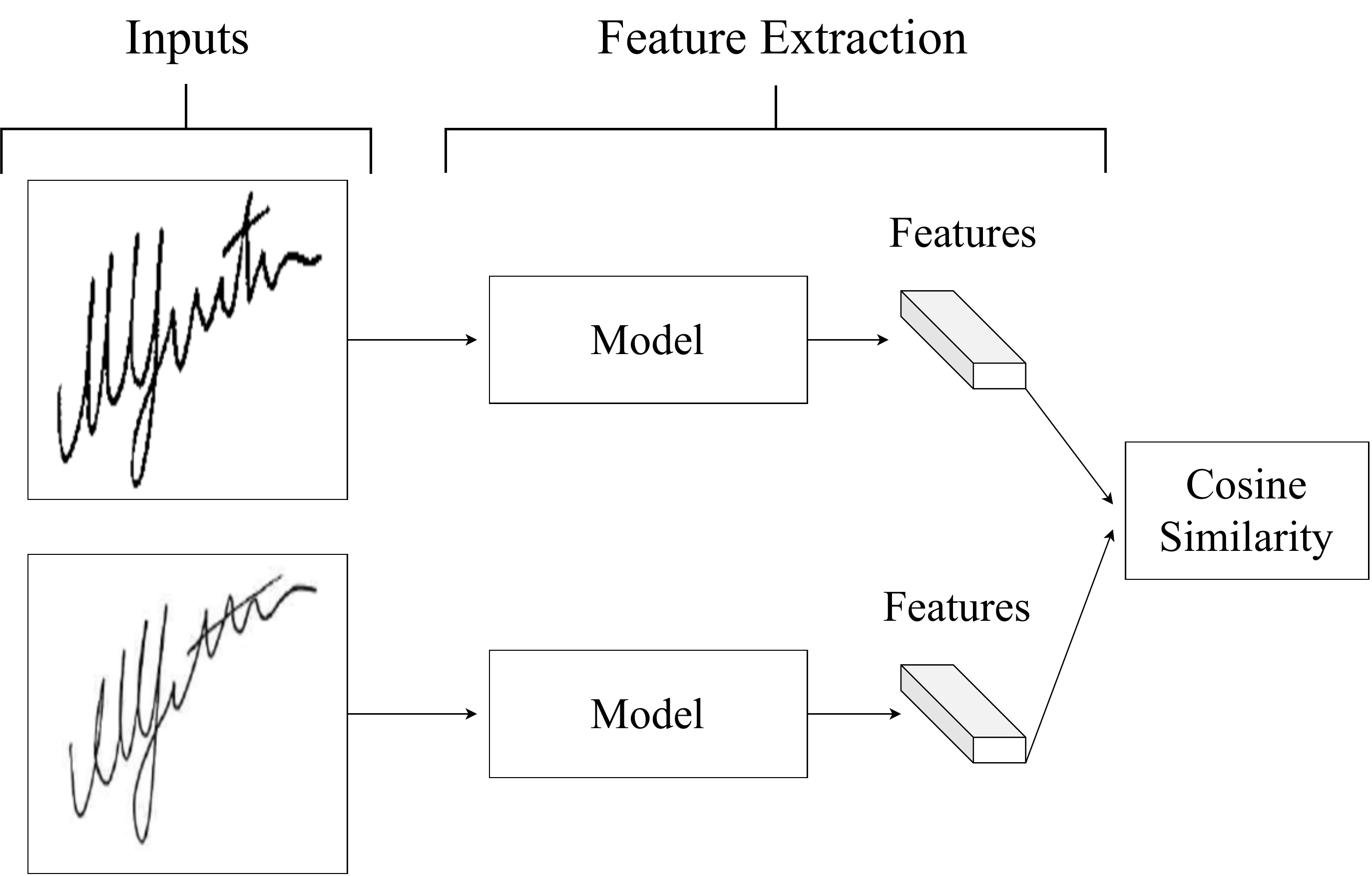}  
  \caption{Feature extraction and verification process.}
  \label{fig:feature_extract}
\end{figure}

Figure~\ref{fig:feature_extract} illustrates the feature extraction and verification process. Two signatures are fed into the model, and their features are extracted. Cosine similarity is calculated between the extracted features. Finally, the obtained similarity score is thresholded to determine whether the signatures belong to the same person or not.

More specifically, the first fully-connected layer of VGG-16 and the second last convolution layer of ResNet-50 are chosen for feature extraction. Accordingly, we obtain a feature vector with size of 4096 from VGG-16 and a feature vector size of 25088 from ResNet-50. Then, we employ cosine similarity to measure the similarity between extracted feature vectors. 
After calculating the similarity for a pair, a label is assigned according to a specified threshold. 

In this paper, we present the results in terms of global equal error rate (EER$_{global}$), based on a global threshold value, and ROC curves. Defining a threshold value for each user is not feasible for a real-world signature verification system, where new users enroll frequently with a few samples provided in a single session.

\section{Experimental Results}

In this section, we first present the datasets and the experimental setups. Then, we will give information about the implementation details. Finally, the objective and subjective evaluation results are provided and discussed.

\subsection{Datasets}

 \begin{figure*}[ht!]
  \centering
\begin{subfigure}{0.19\textwidth}
  \centering
  \includegraphics[width=1\linewidth]{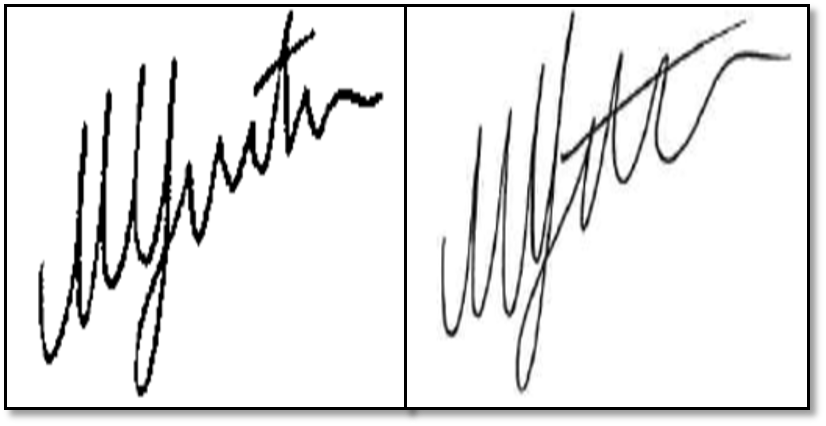}  
  \caption{Setup 1}
  \label{fig:setup1}
\end{subfigure}
\begin{subfigure}{0.19\textwidth}
  \centering
  \includegraphics[width=1\linewidth]{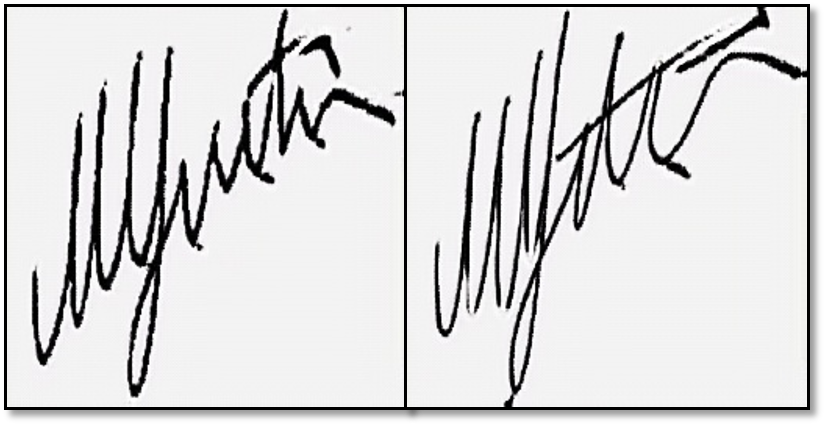}  
  \caption{Setup 2}
  \label{fig:setup2}
\end{subfigure}
\begin{subfigure}{0.19\textwidth}
  \centering
  \includegraphics[width=1\linewidth]{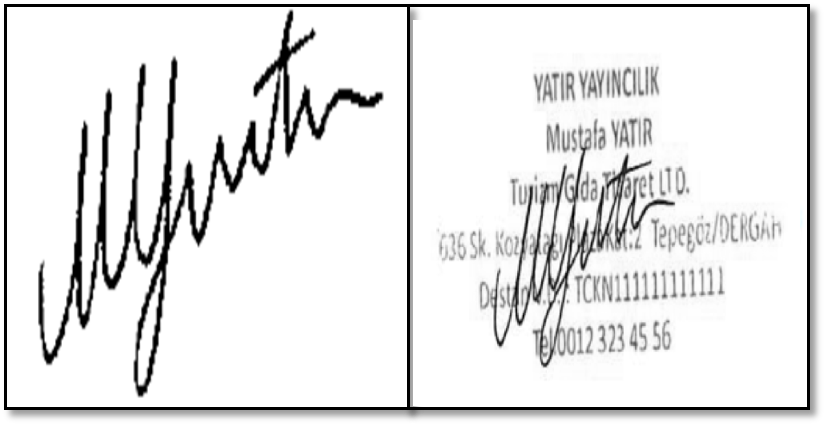}  
  \caption{Setup 3}
  \label{fig:setup3}
\end{subfigure}
\begin{subfigure}{0.19\textwidth}
  \centering
  \includegraphics[width=1\linewidth]{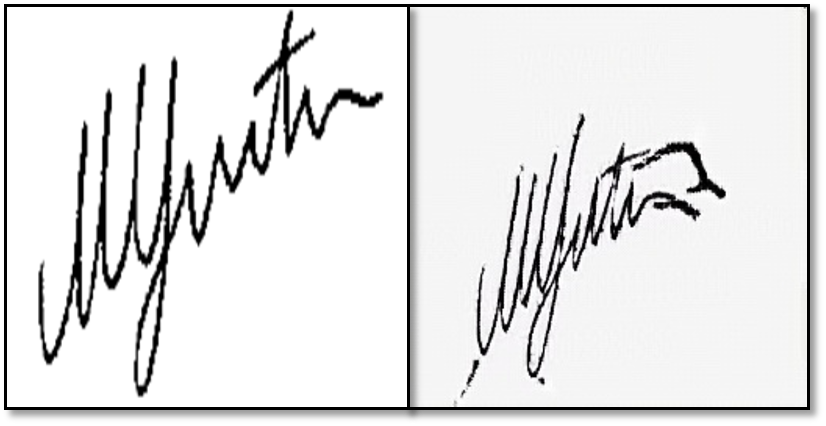}  
  \caption{Setup 4}
  \label{fig:setup4}
\end{subfigure}
\begin{subfigure}{0.19\textwidth}
  \centering
  \includegraphics[width=1\linewidth]{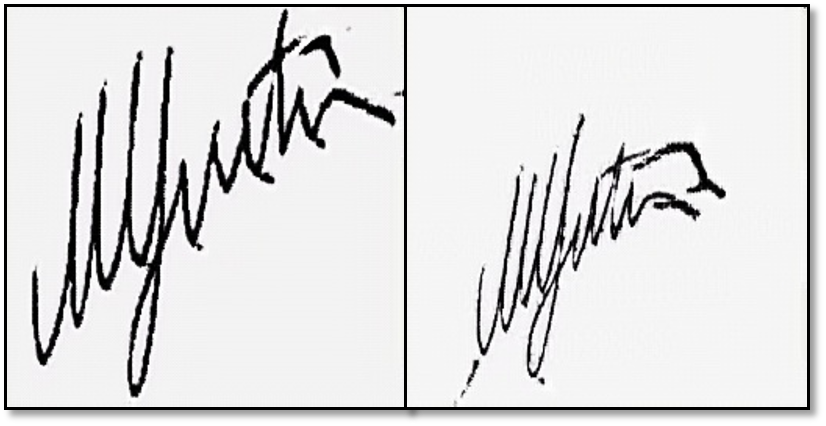}  
  \caption{Setup 5}
  \label{fig:setup5}
\end{subfigure}
\caption{Example pairs of verification test setups. (a) Setup 1: Reference - Unstamped Target, (b) Setup 2: Cleaned Reference - Cleaned Unstamped Target, (c) Setup 3: Reference - Stamped Target, (d) Setup 4: Reference - Cleaned Stamped Target, (e) Setup 5: Cleaned Reference - Cleaned Stamped Target}
\label{fig:verification-setup}
\end{figure*}

We collect signatures from two sources: order documents and signature declaration documents. A sample signature declaration document and an order document can be seen in Figure~\ref{fig:example_sirku_talimat}. \textit{Order documents} include the transaction order of the customers and must be signed by them. Customers must also declare their signatures on \textit{signature declaration documents}. According to the regulations, each person signs three times on signature declaration documents. Signatures extracted from the signature declaration documents are named \textit{reference signatures} of the customers and these are unstamped signatures. On the other hand, signatures extracted from the transaction order documents are named \textit{target signatures}. These signatures can be rubber-stamped or unstamped, which are named as \textit{stamped}, and \textit{unstamped} signatures, respectively. 

\begin{table}[t!]
 \centering
\caption{Verification test setups}
\begin{adjustbox}{width=.47\textwidth}
\centering
\begin{tabular}{@{}l|ll@{}}
\toprule
\textbf{Test setups} & \multicolumn{2}{c}{\textbf{Signature Pairs}}   \\ \midrule
\textbf{Setup 1}     & Reference         & Unstamped Target           \\
\textbf{Setup 2}     & Cleaned Reference & Cleaned Unstamped Target   \\
\textbf{Setup 3}     & Reference         & Stamped Target         \\
\textbf{Setup 4}     & Reference         & Cleaned Stamped Target \\
\textbf{Setup 5}     & Cleaned Reference & Cleaned Stamped Target \\ \bottomrule
\end{tabular}
\label{test-setup}
\end{adjustbox}
\end{table}

Our dataset is categorized into two sub-datasets: (i) representation learning dataset, (ii) verification test dataset. The representation learning dataset is utilized for training a model to learn signature representations. The verification test dataset includes signature pairs (reference signatures and target signatures) to evaluate the signature verification performance. In these datasets, we selected the individuals from whom the bank has received a high number of orders. These two subsets contain different sets of customers, that is, a customer's signatures are included in only one of these two subsets leading to a person independent setup. \\

\textit{\underline{Representation Learning Dataset:}} This dataset consists of 109 individuals' signatures. After applying data augmentation, such as thickening, rotation, and random distortion, each individual has at least 80 signatures. In total, we have approximately 9K signatures. Finally, we split this dataset randomly into training, validation, and test sets with a proportion of 70\%, 15\%, and 15\%, respectively. \\

\textit{\underline{Verification Test Dataset:}} We have two sets of test pairs of signatures from 178 individuals: unstamped pairs and stamped pairs, which consist of reference and target signatures. Unstamped pairs of signatures contain 2609 pairs, which consist of 1001 matched pairs and 1608 mismatched pairs. On the other hand, stamped pairs of signatures contain 2630 pairs, which have 1022 matched pairs and 1608 mismatched pairs. Five different experimental setups are prepared in order to assess the effects of different cases as listed in Table~\ref{test-setup}. Corresponding sample pairs of these setups can be seen in Figure ~\ref{fig:verification-setup}. Please note that the signature images in this figure are resized for visualization purposes. In the first setup, we compare a reference signature with an unstamped signature. In the second setup, we apply our stamp cleaning method both on the reference and unstamped target signature. This is to evaluate the effect of performing a stamp cleaning process when both reference and target signature does not contain any stamps. This could happen, since, at the moment, we do not employ a stamp detection method and apply stamp cleaning on all signatures extracted from the order documents. Stamped target and reference signature are compared in the third setup. This setup is to observe the degree of performance loss when the target signature contains a stamp. In the fourth setup only the stamped target signature is cleaned. This setup is to assess the effect of stamp removal on signature verification performance. Finally, in the fifth setup, both reference and stamped target signatures are cleaned. This setup is to observe the effect of slight artifacts from the cleaning process on the verification performance. Moreover, we could have defined another setup consisting of stamped reference signatures and stamped target signatures. In this case, we should have added a stamp on the reference signature by generating a random stamp; however, the generated stamp cannot be the identical stamp with the target signatures. Since different stamps on the reference and target signatures lead to a decrease in the similarity of these signatures, this is not an appropriate setup for our problem.

\begin{figure}[t!]
  \centering
  \includegraphics[width=1\linewidth]{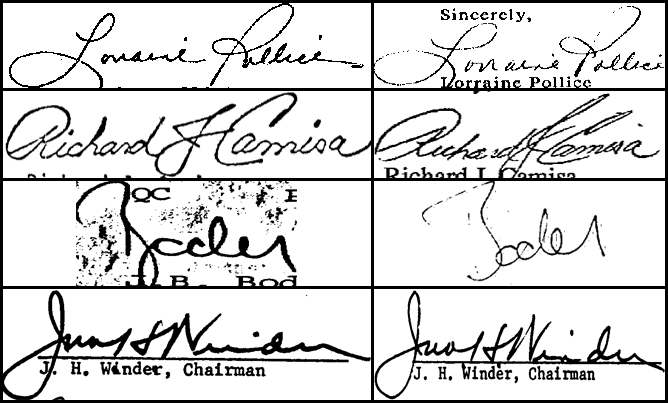} 
\caption{Example signatures of four different individuals from Tobacco-800 dataset.}
\label{fig:tobacco-examples}
\end{figure}

\textbf{Tobacco-800 dataset} \cite{tobacco800,tobacco_website} is a publicly available subset of 42 million pages of documents that are scanned with various equipment. It contains real-world documents and unlike most of the publicly available signature datasets, it contains noises and artifacts, such as stamps, handwritten texts, and ruling lines, on the signatures. Figure \ref{fig:tobacco-examples} shows example signatures of different users from the Tobacco-800 dataset. The resolution of the documents varies between 150 and 300 DPI. All signatures are manually annotated in this dataset. Also, the identification of the users has been done manually by considering the signers' names in the document. There are some mislabeled or unidentified signatures. These mislabeled signatures and signatures without user identities have been removed from the dataset. In the end, 746 signatures of 130 users remained. The number of signatures for each user varies, for example, some users have just one signature. We use randomly selected 60 users to perform representation learning. After applying the same data augmentation strategies with our dataset, we obtain approximately 4200 signatures in total for training.

To perform a writer independent signature verification, we use the remaining 70 users for the test set. 41 of these users only have one signature; therefore, they are only used to generate negative pairs. The remaining 29 users have a minimum of two and a maximum of seven signatures. From these user signatures, we generate all possible positive pairs, which are 166 in the test set. We randomly create the same number of negative pairs by using all the test users. In total, we formed 332 signature pairs. 

\begin{table*}[ht!]
\caption{Signature verification results}
\label{table:verification-all-setups}
\begin{adjustbox}{width=\textwidth}
\begin{tabular}{@{}lcccccc@{}}
\toprule
\multicolumn{1}{c}{\multirow{2}{*}{\textbf{Test Setups}}}                   & \multicolumn{6}{c}{\textbf{EER$_{global}$}}                               \\ \cmidrule(l){2-7} 
\multicolumn{1}{c}{} &
  \textbf{VGG-16} &
  \textbf{VGG-16$_{cleaned}$} &
  \textbf{VGG-16$_{inverse}$} &
  \textbf{ResNet-50} &
  \textbf{ResNet-50$_{cleaned}$} &
  \textbf{ResNet-50$_{inverse}$} \\ \midrule
\textbf{Reference Signature - Unstamped Target Signature}                   & 0.18          & \textbf{0.16} & 0.18          & 0.20 & 0.20 & 0.20 \\
\textbf{Cleaned Reference Signature - Cleaned Unstamped Target Signature}   & 0.18          & \textbf{0.17} & 0.18          & 0.19 & 0.19 & 0.20 \\ \midrule
\textbf{Reference Signature - Stamped Target Signature}                 & 0.33          & \textbf{0.31} & 0.32          & 0.34 & 0.34 & 0.34 \\
\textbf{Reference Signature - Cleaned Stamped Target Signature}         & 0.23          & 0.23          & \textbf{0.22} & 0.26 & 0.26 & 0.26 \\
\textbf{Cleaned Reference Signature - Cleaned Stamped Target Signature} & \textbf{0.22} & \textbf{0.22} & 0.23          & 0.26 & 0.26 & 0.23 \\
\midrule
\textbf{Tobacco-800} & 0.24  & \textbf{0.17}  & -          &  -  &  -  & - \\
\bottomrule
\end{tabular}
\end{adjustbox}
\end{table*}

\begin{figure*}[ht!]
  \centering
\begin{subfigure}{1\textwidth}
  \centering
  \includegraphics[width=1\linewidth]{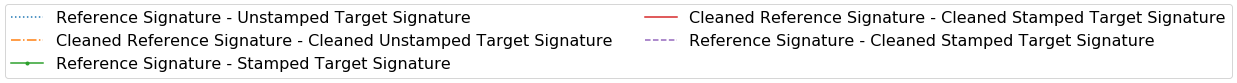}  
  \label{fig:legend}
\end{subfigure}
\begin{subfigure}{.33\textwidth}
  \centering
  \includegraphics[width=1\linewidth]{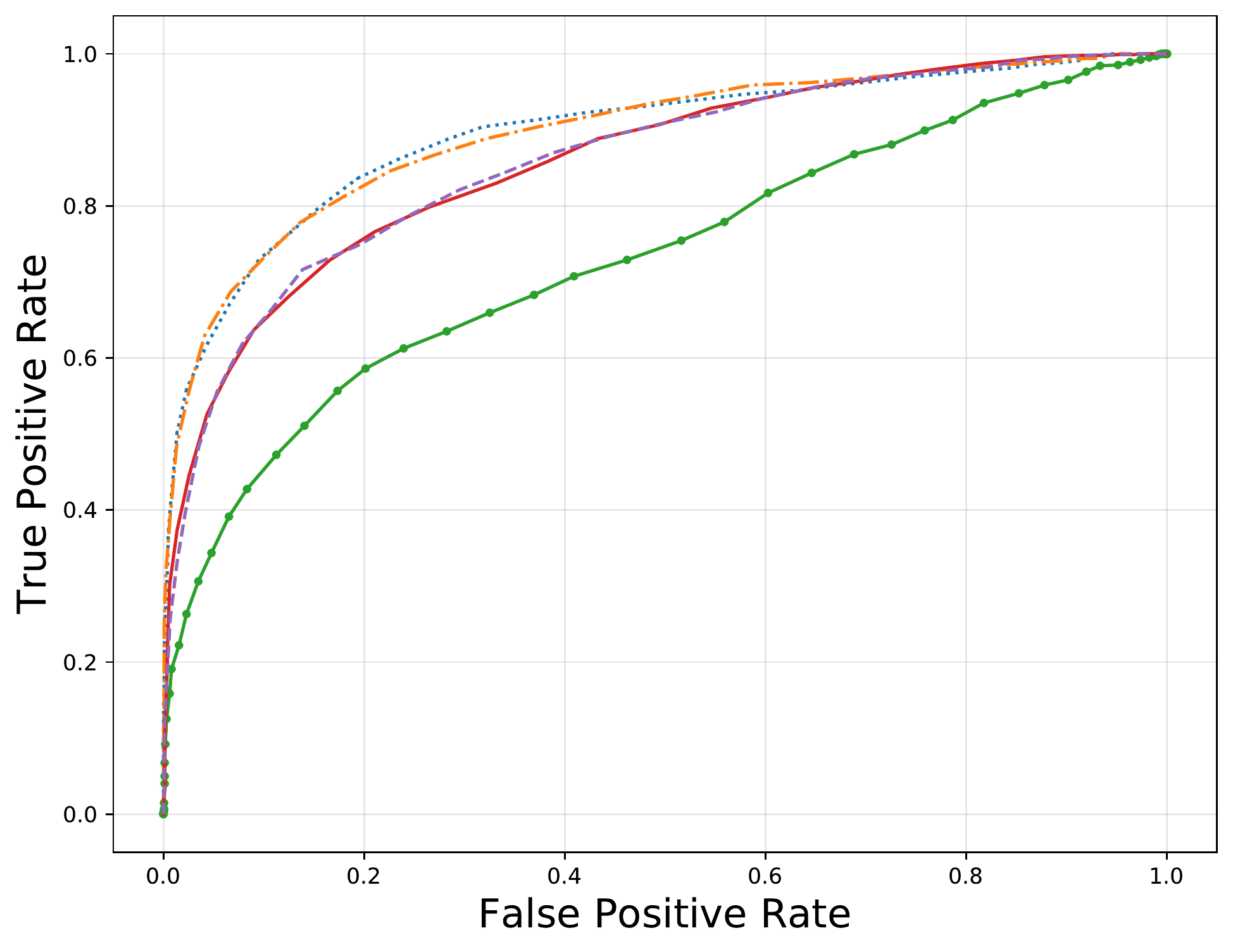}  
  \caption{VGG-16}
  \label{fig:VGG-16}
\end{subfigure}
\begin{subfigure}{.33\textwidth}
  \centering
  \includegraphics[width=1\linewidth]{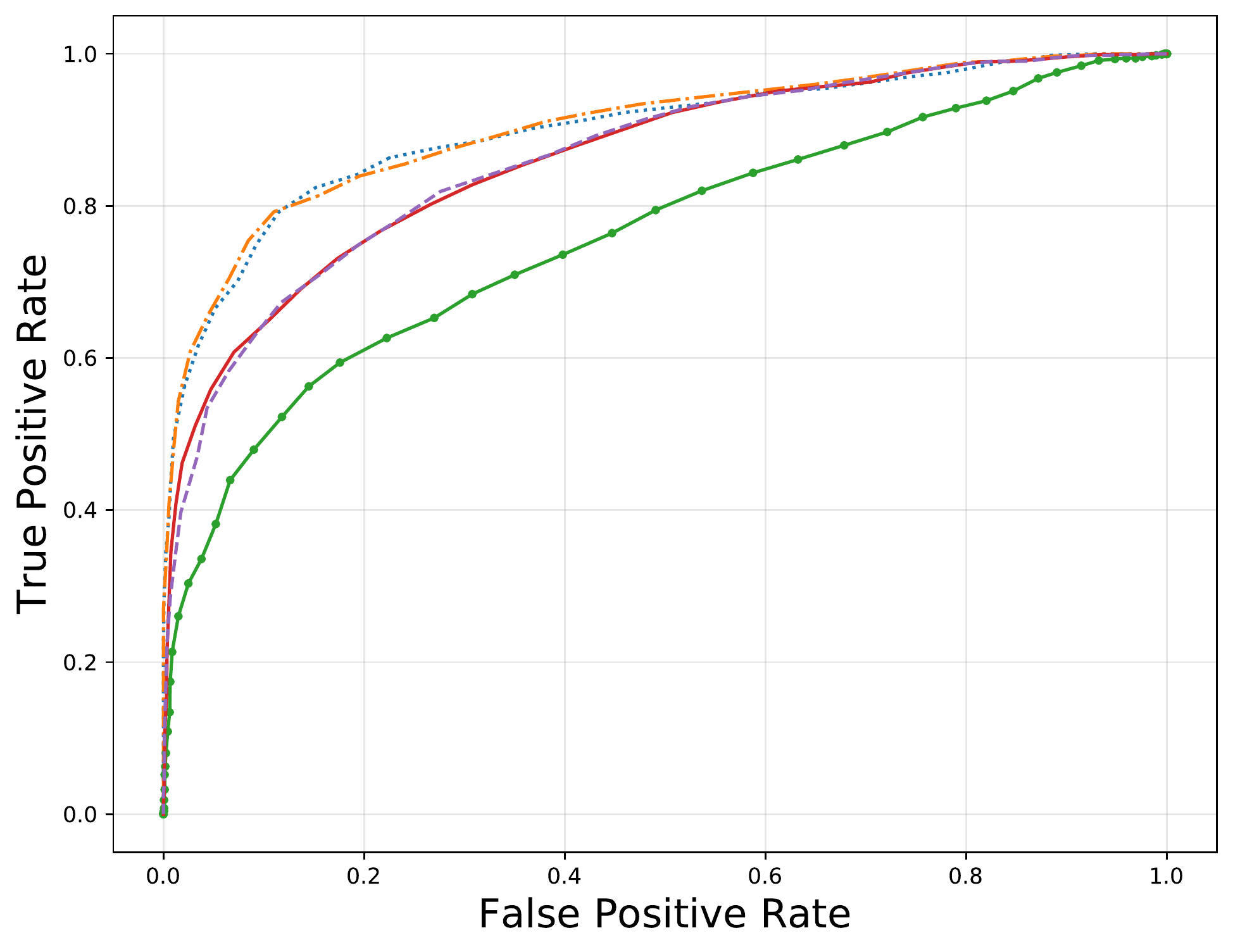}  
  \caption{VGG-16$_{cleaned}$}
  \label{fig:cleaned-VGG-16}
\end{subfigure}
\begin{subfigure}{.33\textwidth}
  \centering
  \includegraphics[width=1\linewidth]{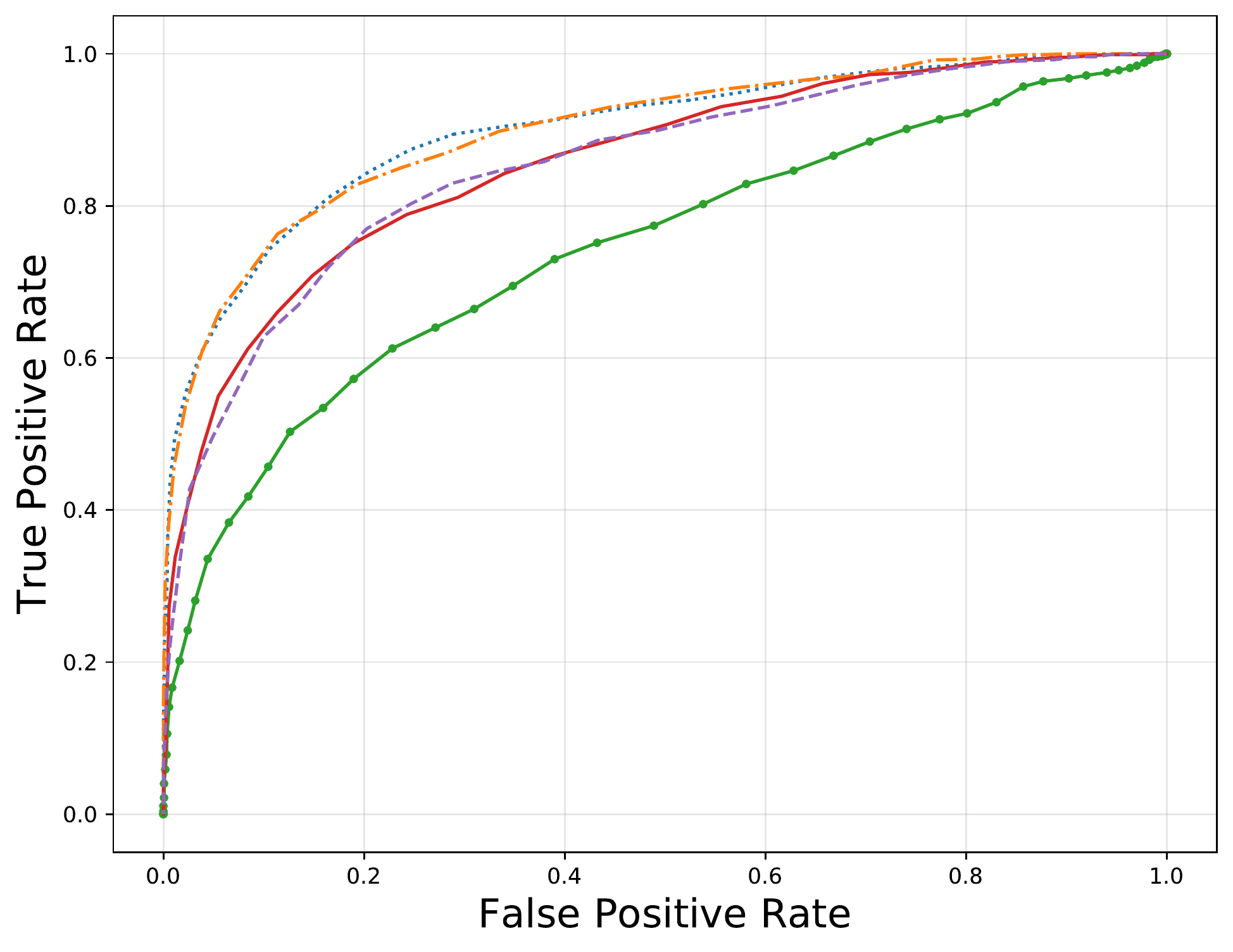}  
  \caption{VGG-16$_{inverse}$}
  \label{fig:inverse-VGG-16}
\end{subfigure}
\begin{subfigure}{.33\textwidth}
  \centering
  \includegraphics[width=1\linewidth]{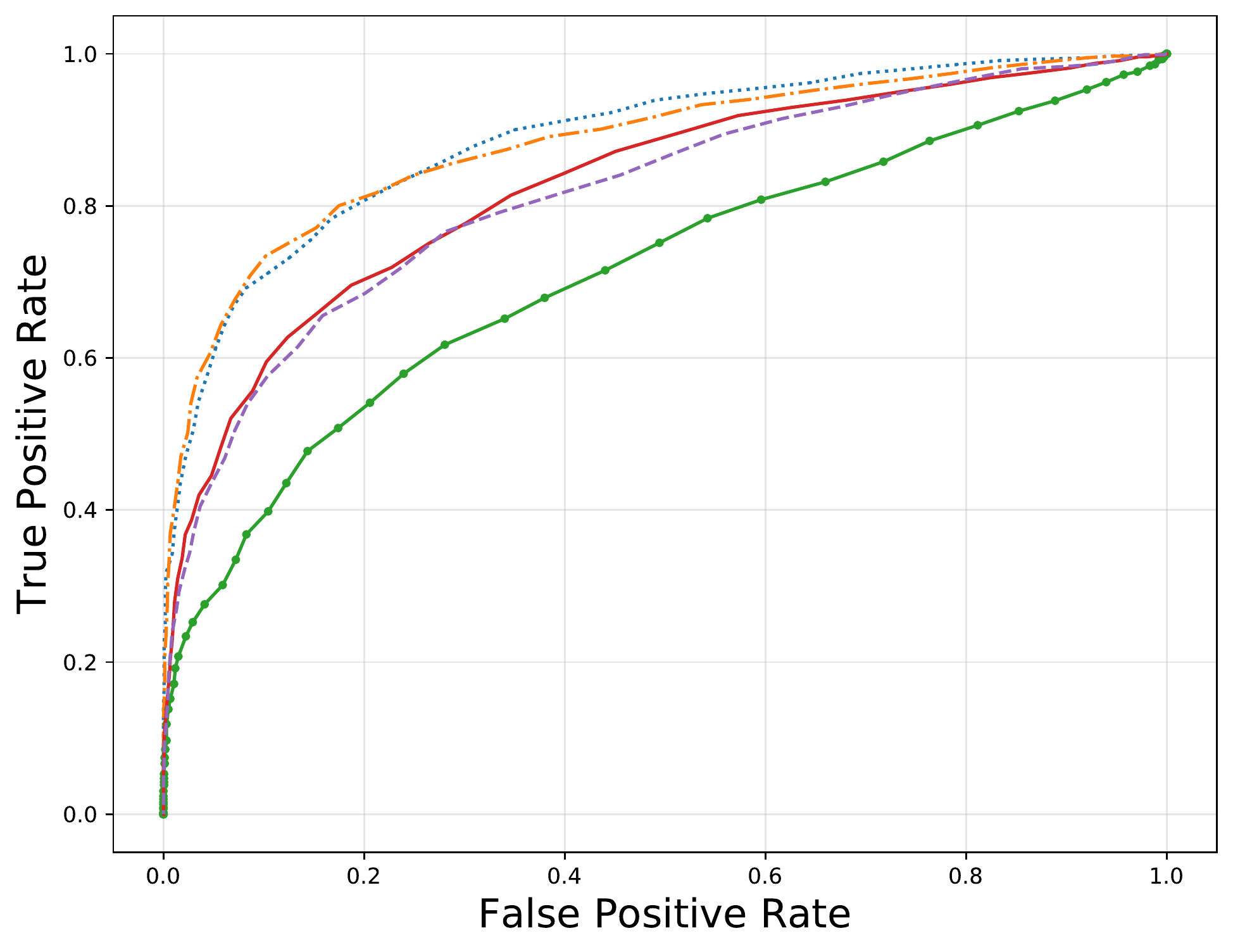}  
  \caption{ResNet-50}
  \label{fig:ResNet-50}
\end{subfigure}
\begin{subfigure}{.33\textwidth}
  \centering
  \includegraphics[width=1\linewidth]{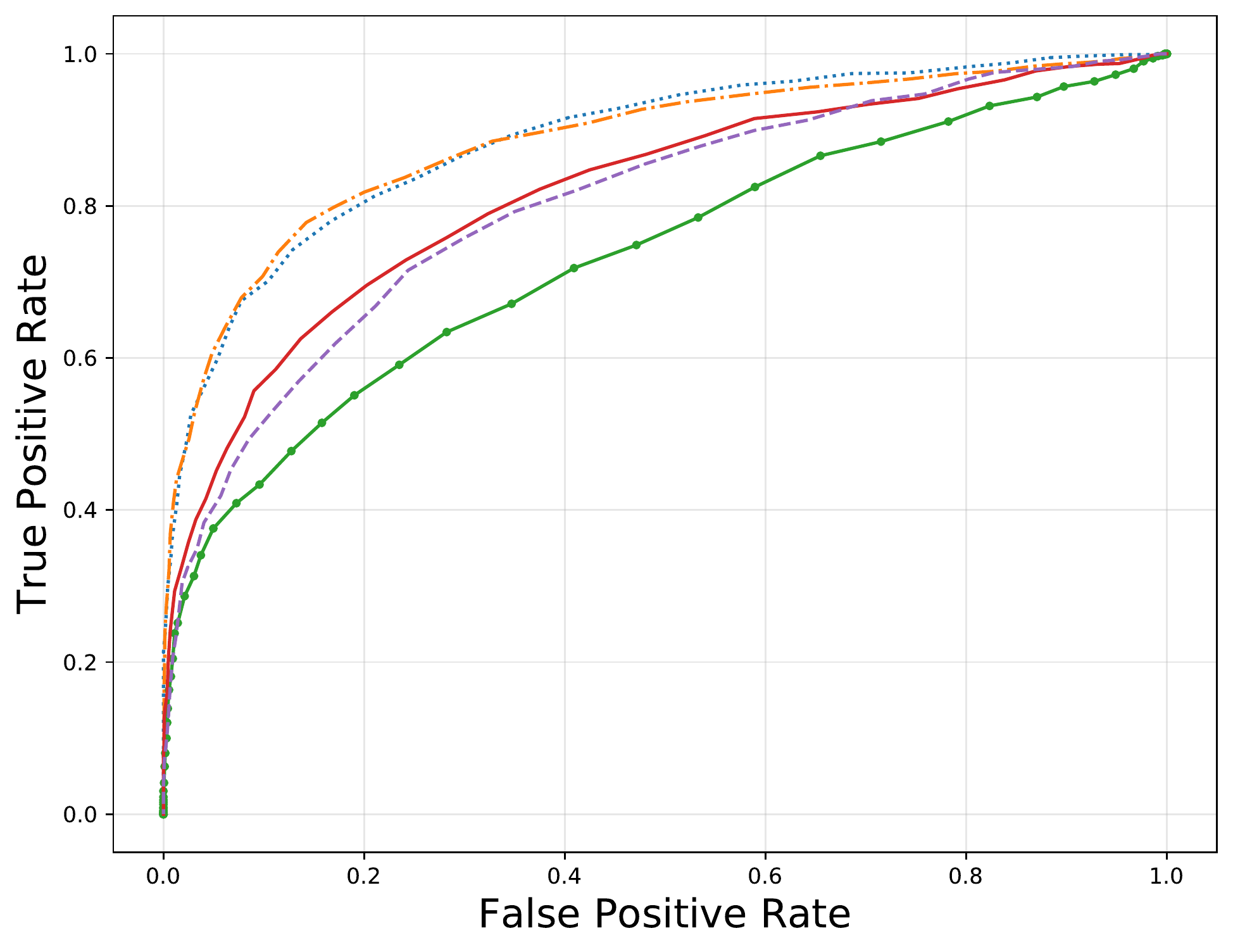}  
  \caption{ResNet-50$_{cleaned}$}
  \label{fig:cleaned-ResNet-50}
\end{subfigure}
\begin{subfigure}{.33\textwidth}
  \centering
  \includegraphics[width=1\linewidth]{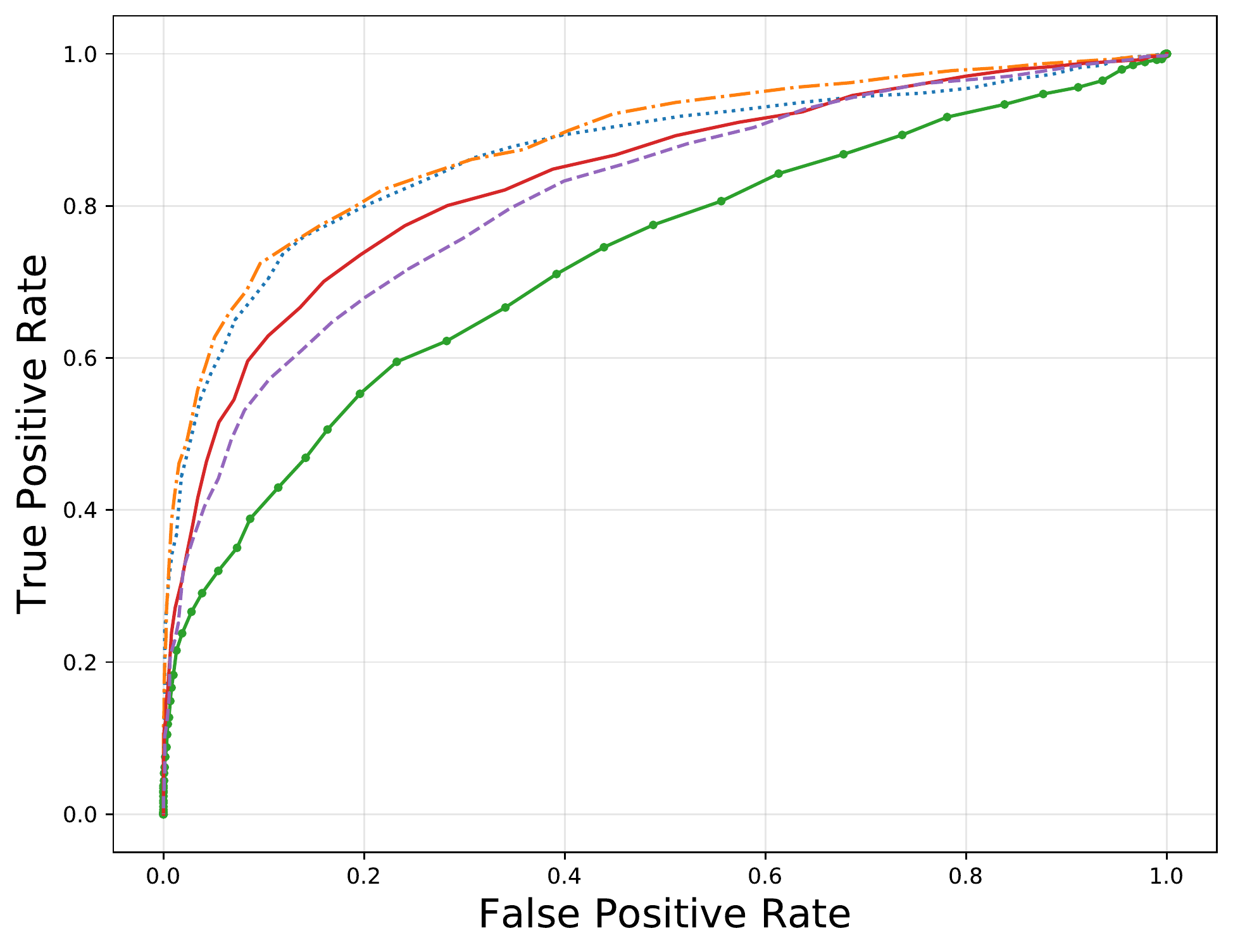}  
  \caption{ResNet-50$_{inverse}$}
  \label{fig:inverse-ResNet-50}
\end{subfigure}
\caption{ROC curves of the model on different experimental setups.}
\label{fig:ROC}
\end{figure*}

\subsection{Implementation Details}
We implement our models in Tensorflow~\cite{tensorflow2015-whitepaper}  and Keras~\cite{chollet2015keras} frameworks. We train our model with NVIDIA GTX 1080Ti graphics card. We perform fine-tuning on ResNet-50 and VGG-16 models with batch size of 32 and 64, respectively. We utilize the SGD optimizer with momentum. The learning rate in the initialization varies in the range of 0.001 and 0.0001. Early stopping is employed by controlling validation loss for specified consecutive epochs.

\subsection{Objective Evaluation}

We run experiments using five different test setups, three different use of fine-tuning data, and three different representations of signature images --original, cleaned, and inverse-- as input. \\

\textbf{Effects of cleaned input images.} We investigate the effectiveness of the stamp cleaning process on signature verification. We train VGG-16 and ResNet-50 on raw input images and cleaned input images, separately. The models trained on the cleaned input images are denoted as VGG-16$_{clean}$ and ResNet50$_{clean}$. We then test these models on five test setups and compare the results. According to Table~\ref{table:verification-all-setups}, the experimental results indicate that the stamps lead to significant degradation of the performance. For example, the obtained EER$_{global}$ with the VGG-16 model is 0.18, when there are no stamps in the target signatures. The EER$_{global}$ increases dramatically to 0.33, when the target signatures contain stamps. However, the cleaning process compensates for this performance loss to a large extent and brings the EER$_{global}$ down to 0.23. This observation is consistent in all the experiments, therefore, independent of the used network model, fine-tuning data, and the input image representation. VGG-16$_{clean}$ model is found to be better than the others in almost all test setups on our dataset. 

ROC curves for all the models are plotted in Figure~\ref{fig:ROC}. Each ROC curve includes the results of five test setups to compare the effects of the cleaning process. As can also be observed from the ROC curves, when \textit{Stamped Target Signatures} are cleaned, the performance increases. When \textit{Unstamped Target Signatures} are cleaned without necessity, the performance does not get affected much. Due to the slight artifacts caused by the cleaning process, applying stamp removal also on the clean reference signature leads to either a slight performance improvement or does not change the performance, depending on the experimental setup. 

We then evaluate our best performing models on the cleaned test pairs of the Tobacco-800 dataset. That is, we train VGG-16 and VGG-16$_{cleaned}$ models on the Tobacco-800 and cleaned Tobacco-800 training sets, respectively. As can be seen from Table~\ref{table:verification-all-setups}, since the Tobacco-800 dataset also consists of real-world documents, the results are  similar to the ones that we have obtained on our custom dataset, which validates the difficulty of the problem.

\textbf{Effects of inverse input images.} For offline signature verification, signature images are digitalized by the scanners. Original images contain a white background and black or blue signatures when scanned. In signature verification literature, we notice that most of the work use binarized signature images with black background and white signatures instead of directly using binarized signature images with white background and black signatures. Therefore, we trained our models with both original and inverse images to see the effect of image representation on the performance. 
From Table~\ref{table:verification-all-setups}, it can be observed that image representation does not affect the verification accuracy significantly. 

\begin{figure*}[h!]
  \centering
  \includegraphics[width=1\linewidth]{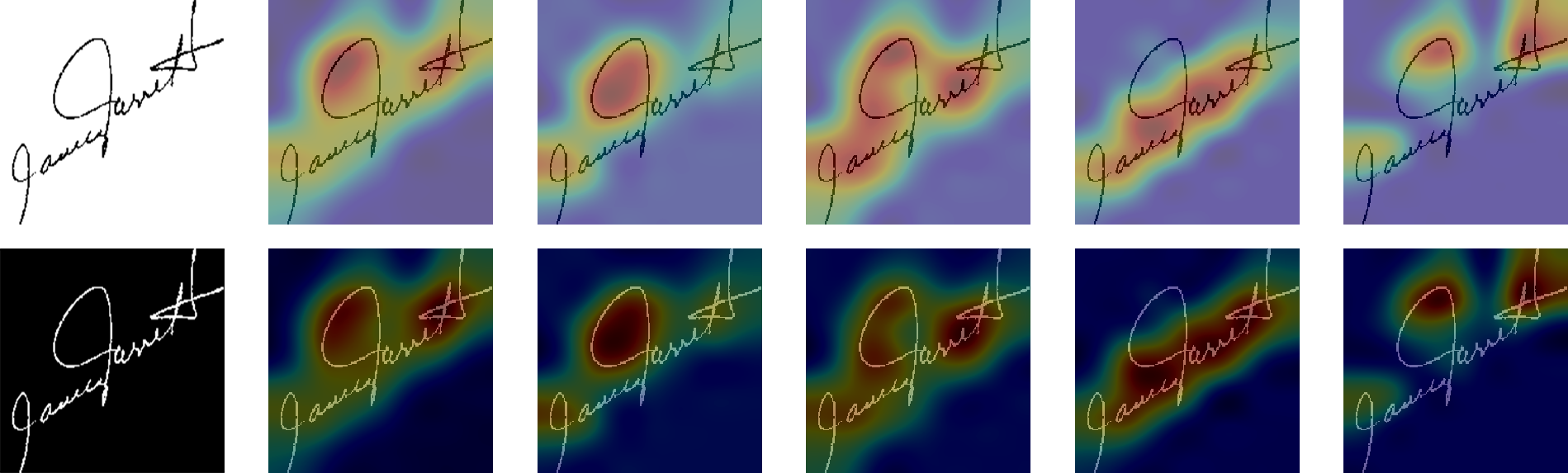}  
  \caption{Response maps with five different filters that have produced highest energy activations in the last convolution layer of VGG-16 network when networks are trained with original signature images where background is white and signature is black (top left) and trained with inverse signature images where background is black and signature is white (bottom left). }
  \label{fig:inversevisualization}
\end{figure*}

To investigate the effect of image representation further, we visualize the five most activated convolution filters of the last convolutional layer for the VGG-16 model. Figure~\ref{fig:inversevisualization} shows that both models, either trained with original or inverse images, learn similar features from the signatures. Visualizations indicate that most activated five convolutional filters concentrate on the same regions of the signatures.

\subsection{Subjective Evaluation}
To assess the difficulty of the problem, we also perform a subjective evaluation by 18 volunteers. We randomly select 360 pairs from our dataset. The subjective evaluation test set includes 180 reference - stamped pairs of signature and 180 reference - unstamped pairs of signatures. These 360 pairs are divided equally into six subsets. Each participant is shown 60 pairs and expected to decide whether the shown signature pair belongs to the same individual or not. This way, each pair is evaluated by three individuals. We provide human evaluation results via majority voting and individual. For majority voting, we assign the human prediction for each pair to whichever prediction is in the majority in the human prediction set. On the other hand, for individual results, we assume having 1080 pairs of signatures and evaluate the prediction of each individual separately.

\begin{table}[h!]
\centering
\caption{Result of subjective evaluation}
\begin{adjustbox}{width=.45\textwidth}
\begin{tabular}{@{}lccc@{}}
\toprule
\multirow{2}{*}{\textbf{Evaluation Method}} & \multicolumn{3}{c}{\textbf{Accuracy (\%)}}                       \\
                                            & \textbf{Human} & \textbf{VGG-16$_{cleaned}$}        & \textbf{ResNet-50$_{cleaned}$}     \\ \midrule
\textbf{Majority Voting}                    & \textbf{91.66} & \multirow{2}{*}{76.38} & \multirow{2}{*}{75.00} \\
\textbf{Individual}                         & \textbf{89.25} &                        &                        \\ \bottomrule
\end{tabular}
\end{adjustbox}
\label{table:AI_human}
\end{table}

In order to compare human vs. machine performance, we also run signature verification experiments with the proposed system on the selected 360 pairs for the subjective evaluation. The models fine-tuned on the cleaned signatures, namely VGG-16$_{cleaned}$ and ResNet-50$_{cleaned}$, are chosen to extract features. EER$_{global}$ is calculated on these pairs, and the threshold value according to this EER$_{global}$ is used to calculate the accuracy of the models. 

Results of human evaluation, along with the accuracies obtained by the models, are given in Table~\ref{table:AI_human}. The results show the challenging nature of the task  as even humans cannot predict all the pairs correctly. 
Model accuracies on this subset are lower than the ones obtained on the overall test set in Table~\ref{table:verification-all-setups}, which indicates that the chosen subset includes harder pairs. Comparing human and model performances, it is clear that we still need further improvements in the system to match human performance.

\section{Conclusion}

In this paper, we have presented a comprehensive study on writer independent offline signature verification in a real-world scenario, where occluded signatures of a bank's customers' are verified against their clean reference signatures. We have proposed a CycleGAN based stamp removal method to clean signatures before feeding them to a CNN model to extract the signature representation. We have compared different verification setups, fine-tuning strategies, and signature representation approaches and analyzed their effects. In order to show the difficulty of the problem, we have also conducted a human evaluation. We have shown the challenging nature of the problem and effectiveness of our proposed stamp cleaning method in our experiments both on our custom dataset and on publicly available Tobacco-800 dataset.\\



\noindent\textbf{Acknowledgements.} We would like to thank our colleagues from Applied AI and R\&D Department of Yap{\i} Kredi Technology, especially Ali Ye{\c{s}}ilkanat and Mehmet Yasin Akp{\i}nar for their support and valuable comments.

{\small
\bibliographystyle{ieee_fullname}
\bibliography{egbib}
}

\end{document}